\documentclass[conference]{IEEEtran}
\IEEEoverridecommandlockouts
\usepackage{cite}
\usepackage{amsmath,amssymb,amsfonts}
\usepackage{algorithmic}
\usepackage{graphicx}
\usepackage{textcomp}
\usepackage{xcolor}
\def\BibTeX{{\rm B\kern-.05em{\sc i\kern-.025em b}\kern-.08em
    T\kern-.1667em\lower.7ex\hbox{E}\kern-.125emX}}

\usepackage[ruled,vlined]{algorithm2e}
\usepackage{flushend}

\usepackage{bm}
\usepackage{subfig}
\usepackage{booktabs}
\usepackage{indentfirst}
\usepackage{multicol}
\usepackage{multirow}
\usepackage{epsfig}
\usepackage[normalem]{ulem}
\usepackage{bm}

\begin{document}

\title{Fine-grained Domain Adaptive Crowd Counting \\
via Point-derived Segmentation \\
\thanks{*Shengfeng He is the corresponding author (shengfenghe7@gmail.com)}
}

\author{
Yongtuo Liu\textsuperscript{1,2}, Dan Xu\textsuperscript{3}, Sucheng Ren\textsuperscript{1}, Hanjie Wu\textsuperscript{1}, Hongmin Cai\textsuperscript{1}, Shengfeng He\textsuperscript{1,4*}\\\\
\textsuperscript{1}School of Computer Science, South China University of Technology\\
\textsuperscript{2}Institute of Informatics, University of Amsterdam\\
\textsuperscript{3}Department of Computer Science and Engineering, Hong Kong University of Science and Technology\\
\textsuperscript{4}School of Computing and Information Systems, Singapore Management University
}

\maketitle

\begin{abstract}
Due to domain shift, a large performance drop is usually observed when a trained crowd counting model is deployed in the wild.
While existing domain-adaptive crowd counting methods achieve promising results, they typically regard each crowd image as a whole and reduce domain discrepancies in a holistic manner, thus limiting further improvement of domain adaptation performance.
To this end, we propose to untangle \emph{domain-invariant} crowd and \emph{domain-specific} background from crowd images and design a fine-grained domain adaption method for crowd counting.
Specifically, to disentangle crowd from background, we propose to learn crowd segmentation from point-level crowd counting annotations in a weakly-supervised manner.
Based on the derived segmentation, we design a crowd-aware domain adaptation mechanism consisting of two crowd-aware adaptation modules, i.e., Crowd Region Transfer (CRT) and Crowd Density Alignment (CDA).
The CRT module is designed to guide crowd features transfer across domains beyond background distractions.
The CDA module dedicates to regularising target-domain
crowd density generation by its own crowd density distribution.
Our method outperforms previous approaches consistently in the widely-used adaptation scenarios.
\end{abstract}

\begin{IEEEkeywords}
Crowd Counting, Domain Adaptation, Point-derived Segmentation
\end{IEEEkeywords}

\section{Introduction}

Crowd counting has drawn increasing attention because of its fundamental role in social management~\cite{li2014crowded, gao2020cnn}.
Due to domain shift \cite{torralba2011unbiased}, performance usually degrades a lot when trained crowd counting models are deployed in unseen crowd scenes.
To fill the performance gap, a direct solution is to massively label abundant images in each crowd scene.
However, the labeling is quite onerous for crowd counting as it requires labeling all human heads in each crowd image.

\begin{figure}[t]
 \centering
 \setlength{\tabcolsep}{1pt}
 \includegraphics[width=.98\linewidth]{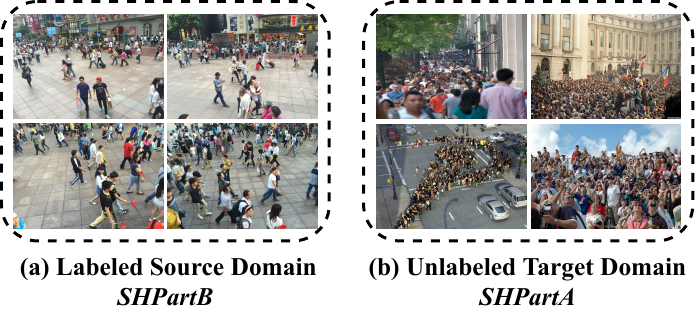}
 \vspace{-8pt}
 \caption{Crowd images from \emph{SHPartB} (a) and \emph{SHPartA} (b) datasets~\cite{zhang2016single} respectively. As can be seen, backgrounds vary a lot across domains. For instance, backgrounds in \emph{SHPartB} are mainly ground whereas various backgrounds appear in \emph{SHPartA} including buildings, trees, sky, etc.}
\vspace{-8pt}
 \label{teaser}
\end{figure}

To avoid labeling burdensome, one promising way is to introduce Unsupervised Domain Adaptation (UDA) to transfer essential knowledge learned from a labeled source domain to a related but unlabeled target domain~\cite{pan2009survey}.
Recently, several methods are proposed to apply UDA for domain-adaptive crowd counting, including pixel-level adaptation methods~\cite{wang2019learning, gao2021domain} and feature-level adaptation methods~\cite{hossain2020domain, han2020focus, gao2020feature, zou2021coarse}.
The feature-level methods can achieve competitive performance and work efficiently, thus dominating the existing literature.

While achieving promising results, existing domain adaptative crowd counting methods reduce domain discrepancies on crowd and background simultaneously.
The holistic manner inevitably degrades domain adaptation performance considering that domain-specific background varies a lot across domains (shown in Fig.~\ref{teaser}) and background alignment across domains challenges domain-invariant representation learning, which further harms the discrimination of crowd and background critical for crowd counting~\cite{wang2020classes, wang2020progressive}.

To this end, we propose to treat crowd and background differently while conducting domain adaptation.
Note that crowd counting only labels one point per human without segmentation. To untangle crowd and background from point-level annotations, we learn crowd segmentation from the sparse point annotations in a weakly-supervised manner.
Based on the derived segmentation, we propose a Crowd-aware domain Adaptation framework for Crowd Counting (CACC), which consists of two crowd-aware adaptation modules, namely Crowd Region Transfer (CRT) and Crowd Density Alignment (CDA).
Specifically, to guide crowd alignment across domains beyond background distractions, we introduce the CRT module to bridge domains by learning domain-invariant crowd features.
Besides, we introduce the CDA module to generate segmentation-guided pseudo labels in the target domain to regularize crowd density generation by target-domain's own crowd density distribution, instead of by source-domain crowd density distribution utilized in the previous methods~\cite{ han2020focus, gao2020feature, zou2021coarse}. The design considers that different domains usually have quite different crowd density distributions, as shown in Fig.~\ref{teaser}.
It inevitably degrades the adaptation performance to directly utilize source-domain crowd density labels to regularize target-domain crowd density distribution.

In summary, the contributions are organized as follows:
\begin{itemize}
\item
We propose to treat crowd and background differently and design a crowd-aware mechanism for domain adaptive crowd counting. 
\item
We propose a simple and effective schema to derive segmentation from point-level crowd counting annotations.
Two crowd-aware domain adaptation modules are further proposed, based on point-derived segmentation, to guide crowd features transfer across domains beyond background distraction and regularize target-domain crowd density generation.
\item
Our method outperforms previous approaches consistently in the widely-used adaptation scenarios.
\end{itemize} 

\section{Related Work}

\noindent \textbf{Domain-adaptive Crowd Counting.}
Recently, some methods are proposed to solve domain-adaptive crowd counting.
They can be mainly grouped into three categories.
(i) \emph{Pixel-level} adaptation methods~\cite{wang2019learning}: \cite{wang2019learning} constructs a synthetic dataset GCC and modifies CycleGAN~\cite{zhu2017unpaired} to conduct style transfer to generate target-domain crowd images for supervised training.
(ii) \emph{Feature-level} adaptation methods: Gao \emph{et al.}~\cite{gao2020feature} propose to discriminate features across domains and constrain density map generation by source-domain density labels.
Han \emph{et al.}~\cite{han2020focus} constrain the feature extraction by a feature discriminator and an auxiliary semantic task.
Hossain \emph{et al.}~\cite{hossain2020domain} reduce the domain shift by minimizing the feature distances (i.e., Maximum Mean Discrepancy (MMD)~\cite{long2017deep}) across domains.
(iii) \emph{Others}~\cite{liu2020towards, wang2021neuron, wu2021dynamic, he2021error}: \cite{liu2020towards} introduce an extra head detector for mutual training with the crowd counter. \cite{wang2021neuron} present a neuron linear transformation to optimize a small amount of parameters based on a few target-domain training samples.
\cite{wu2021dynamic} introduce an external template encoding domain-specific meta information for humans.
\cite{he2021error} exploit a density isomorphism reconstruction objective
derived from consecutive frames in crowd videos.
Methods in \emph{others} can be regarded as supplements  with additional bounding box annotations~\cite{liu2020towards}, extra target-domain annotations~\cite{wang2021neuron}, an external template encoding~\cite{wu2021dynamic}, or temporal consistency in videos~\cite{he2021error}.

\begin{figure*}[t]
 \centering
 \setlength{\tabcolsep}{1pt}
 \includegraphics[width=.8\linewidth, height=6.4cm]{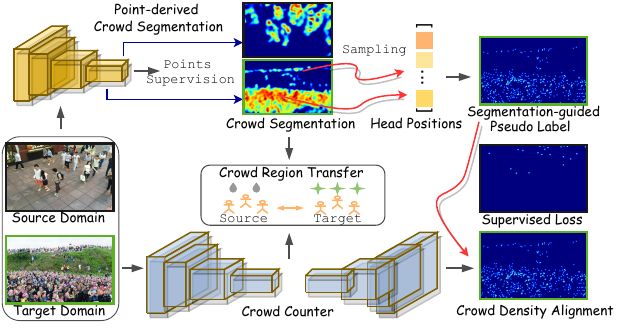}
 \vspace{-10pt}
 \caption{Overview of the proposed Crowd-aware domain Adaptation framework for Crowd Counting (CACC). To disentangle crowd from background, we derive crowd segmentation from point-level crowd counting annotations, namely Point-derived Crowd Segmentation (PCS), in a weakly-supervised manner. Based on the derived segmentation, we propose two crowd-aware adaptation modules, i.e., Crowd Region Transfer (CRT) and Crowd Density Alignment (CDA). Crowd Region Transfer guides crowd features alignment across domains beyond background distractions. Crowd Density Alignment samples pseudo head positions from segmentation to generate segmentation-guided pseudo labels, which are utilized to regularize target-domain crowd density generation by its own crowd density distribution.}
 \vspace{-10pt}
 \label{network architecture}
 \end{figure*}

While effective, they all conduct domain adaptation in a holistic manner.
However, domain alignment between crowd and background inevitably incurs misalignment, leaving room for improvement of previous methods.

\noindent \textbf{Crowd Counting and Domain Adaptation.} Due to limited space, generic crowd counting and domain adaptation methods are discussed in the Appendix. 

\section{Method}
\vspace{-4pt}
\subsection{Problem Formulation}
\vspace{-6pt}
In domain adaptive crowd counting, we are given a labeled source domain $\mathcal{D_{S}}=\{(\mathbf{x}_{i}^{s}, \mathbf{y}_{i}^{s})\}_{i=1}^{N_{s}}$ where $\mathbf{x}_{i}^{s}$ and $\mathbf{y}_{i}^{s}$ denote the $i$-th crowd image and the corresponding annotation, i.e., coordinates of head positions. Besides, we have access to a unlabeled target domain $\mathcal{D_{T}}=\{(\mathbf{x}_{i}^{t})\}_{i=1}^{N_{t}}$.
Our goal is to improve counting performance in the unlabeled target domain $\mathcal{D_{T}}$ utilizing knowledge from both domains.

\vspace{-3pt}
\subsection{Framework Overview}
\vspace{-5pt}
As shown in Fig.~\ref{network architecture}, we propose a Crowd-aware domain Adaptation framework for Crowd Counting (CACC), which contains a crowd counter, a Point-derived Crowd Segmentation (PCS) network, and two crowd-aware adaptation modules, i.e., Crowd Region Transfer (CRT) and Crowd Density Alignment (CDA). Details of the basic crowd counter are in the Appendix.

\vspace{-3pt}
\subsection{Point-derived Crowd Segmentation}
\vspace{-5pt}
Point-derived Crowd Segmentation (PCS) is proposed to disentangle crowd from background by point-level crowd counting annotations in a weakly-supervised manner.
The rationale behind this design is that although point annotations do not specify segmentation, they still entail where crowd appears and how crowd looks from a statistical perspective. This is also studied in the context of Multiple Instance Learning (MIL)~\cite{maron1998framework} where a label is assigned to each bag of instances instead of each instance. In our case, each patch cropped from crowd images can be regarded as a bag of pixels where patch-level labels can be defined as follows. 

Specifically, we densely sample patches from crowd images to construct crowd or background bags $\mathcal{B} = \{\mathbf{b}_1, \mathbf{b}_2, ..., \mathbf{b}_N\}$.
Each patch $\mathbf{b}_i$ in $\mathcal{B}$ contains a set of pixels $\mathbf{X}_i = \{x_1, x_2, ..., x_{h_i \times w_i}\}$.
Let $y_j$ be the label of each pixel $x_j$ which indicates whether it is annotated in crowd counting.
Following the standard MIL assumption that a negative bag contains only negative instances while a positive bag contains at least one positive instance, we partition $\mathcal{B}$ into crowd bags $\mathcal{B}_C$ and background bags $\mathcal{B}_B$ according to whether a bag contains at least a crowd counting annotation or not:
\begin{equation}
\label{PCS_BC_BB}
\begin{split}
\mathcal{B}_C = \{\mathbf{b}_i \in \mathcal{B}\ if\ y_j = 1, \exists\ x_j \in \mathbf{b}_i\},  \\
\mathcal{B}_B = \{\mathbf{b}_i \in \mathcal{B}\ if\ y_j = 0, \forall\ x_j \in \mathbf{b}_i\}.
\end{split}
\end{equation}


\begin{figure}[t]
 \begin{center}
 \setlength{\tabcolsep}{1.0pt}
 \renewcommand{\arraystretch}{0.8}
 \includegraphics[width=.99\linewidth, height=2.8cm]{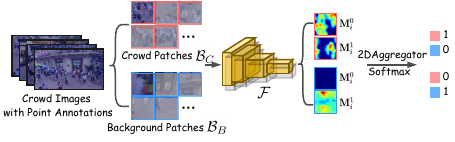}
\end{center}
\vspace{-14pt}
\caption{Learning Point-derived Crowd Segmentation (PCS) from point-level crowd counting annotations. 0 and 1 in each side constitute a one-hot vector indicating the patch is annotated as a crowd or background patch.}
\vspace{-15pt}
\label{PCS}
\end{figure}

To learn segmentation from patch-level labels, we build a learner $\mathcal{F}$ which classifies crowd and background patches.
Given each sample $\mathbf{b}_i$ from $\mathcal{B}_C$ or $\mathcal{B}_B$, $\mathcal{F}$ outputs an intermediate 2-channel map $\mathbf{M}_i = \mathcal{F}(\mathbf{b}_i,\Theta)$.
Optimization objective of classifier $\mathcal{F}$ is a standard cross entropy loss:
\begin{equation}
\label{learnF}
\begin{split}
  \mathcal{L_F} = \sum_{\mathbf{b}_{i} \in \mathcal{B}_C}-\log(\mathcal{S}(\mathcal{A}(\mathbf{M}_i^0)))
            \\    +
                \sum_{\mathbf{b}_{i} \in \mathcal{B}_B}-\log(\mathcal{S}(\mathcal{A}(\mathbf{M}_i^1))),
\end{split}
\end{equation}
where $\mathcal{A}(\cdot)$ is a 2D aggregator (e.g., Avg2D), $\mathcal{S}(\cdot)$ is the softmax function. $\mathbf{M}_i^0$ and $\mathbf{M}_i^1$ represent the first and second channel of $\mathbf{M}_i$.
The learning of $\mathcal{F}$ can activate pixel-wise responses in $\mathbf{M}_i$ for better discrimination of crowd and background patches.
After learning converges, we utilize $\mathbf{M}_i$ as crowd and background segmentation. PCS is shown in Fig.~\ref{PCS}.

Note that crowd counting annotates human head positions only. A background bag $\mathcal{B}_B$ in Eq.~(\ref{PCS_BC_BB}) may contain human bodies. However, $\mathcal{F}$ in Eq. (\ref{learnF}) is trained statistically with natural tolerance to noisy labels. 
As shown in Fig.~\ref{network architecture}, we can still derive high-quality crowd segmentation from noisy labels.
As human heads only exist in crowd bags $\mathcal{B}_C$, activated crowd segmentation can thus focus more on human heads for better classification.
This is a blessing in disguise when head-highlighted crowd segmentation is utilized to design crowd-aware domain adaptation modules. This is because of the head-centric labeling and recognition nature of crowd counting. We will detail the benefits of head-highlighted crowd segmentation in the following proposed modules.

 In practice, only the source domain has crowd counting annotations. We utilize density maps estimated by crowd counter for training of $\mathcal{F}$ in the target domain. As discussed above, $\mathcal{F}$ does not rely on accurate labels due to its statistical nature. As shown in Fig.~\ref{PCS samples}, segmentation results in the target domain are still ensured high quality.
 
\vspace{-4pt}
\subsection{Crowd Region Transfer}
\vspace{-4pt}
Crowd Region Transfer (CRT) is designed to align crowd features across domains beyond background distractions by learning domain-invariant crowd feature representations.


Given a crowd image $\mathbf{x}$, we denote crowd segmentation from Point-derived Crowd Segmentation as $\mathbf{C}_{seg}$.
We have two variants to design segmentation, i.e., soft crowd segmentation $\mathbf{C}_{seg}^S$ and hard crowd segmentation $\mathbf{C}_{seg}^H$.
We directly utilize $\mathbf{C}_{seg}$ as $\mathbf{C}_{seg}^S$.
For $\mathbf{C}_{seg}^H$, we binarize $\mathbf{C}_{seg}^S$ by:
\begin{equation}
\label{HNCS}
T = \frac{1}{HW}\sum_{h,w}\mathbf{C}_{seg}^{S}(h,w), \,\,
\mathbf{C}_{seg}^H = \mathcal{I}(\mathbf{C}_{seg}^S > T),
\end{equation}
where threshold $T$ is set to the mean value of $\mathbf{C}_{seg}^S$, $\mathcal{I}(\cdot)$ represents the indication function.

Following~\cite{han2020focus, gao2020feature, zou2021coarse}, we utilize adversarial training to learn domain-invariant features. Differently, we discard domain-specific background features and focus on domain-invariant crowd features.
Optimization objective of CRT is:
\begin{equation}
\label{CRT part loss function}
\begin{split}
\mathcal{L}_{C\!RT} =\,\, & \underset{\theta_{G}}{{\rm min}} \, \underset{\theta_{D}}{{\rm max}} \, \mathbb{E}_{\mathbf{x}_s\sim\mathcal{D_S}} \, {\rm log} \, D(G(\mathbf{x}_s)\cdot\mathbf{C}_{seg}^{H/S}) \\
&+ \mathbb{E}_{\mathbf{x}_t\sim\mathcal{D_T}} \, {\rm log}(1 - D(G(\mathbf{x}_t)\cdot\mathbf{C}_{seg}^{H/S})),
\end{split}
\end{equation}
where $G$ is the feature extractor of crowd counter. $D$ is domain classifier.
$D$ and $G$ construct a two-player minimax game, where $D$ is trained to distinguish which domain the features come from, while $G$ aims to confuse $D$.

Note that soft crowd segmentation $\mathbf{C}_{seg}^{S}$ from PCS is head-highlighted. When utilized in Eq.~\ref{CRT part loss function}, $\mathbf{C}_{seg}^{S}$ enhances head features alignment across domains, which is crucial for crowd counting due to its head-centric recognition mechanism. Effectiveness of $\mathbf{C}_{seg}^{S}$ is shown in Table~\ref{CRT Table}.


\begin{algorithm}[t]
\caption{Crowd-aware Domain Adaptation for Crowd Counting.}
\label{optimization procedure}
\LinesNumbered 
\KwIn{Labeled source domain $\mathcal{D_{S}}$. \qquad \qquad \qquad \qquad
      Unlabeled target domain $\mathcal{D_{T}}$. \qquad \qquad \qquad \qquad \qquad \qquad \qquad \qquad \qquad \qquad \qquad \qquad Batch size $B$.}
\KwOut{A domain adaptive crowd counter $C(\cdot,\theta)$.}
Supervised learning of $C(\cdot,\theta)$ in $\mathcal{D_{S}}$.\\
Sample bags $\mathcal{B}$ in $\mathcal{D_{S}}$ \& $\mathcal{D_{T}}$ and partition $\mathcal{B}$ into $\mathcal{B}_C$ and $\mathcal{B}_B$ by Eq.~(\ref{PCS_BC_BB}).\\
Learning of $\mathcal{F}$ on $\mathcal{B}_C$ and $\mathcal{B}_B$ by Eq.~(\ref{learnF}).\\
Obtain crowd segmentation in $\mathcal{D_{S}}$ and $\mathcal{D_{T}}$.\\
\For{$i = 1$ {\rm \textbf{to}} $max\_iter$}{
    $X_S, Y_S \leftarrow Sample(\mathcal{D_{S}}, B/2)$\\
    $X_T \leftarrow Sample(\mathcal{D_{T}}, B/2)$\\
    Calculate $\mathcal{L}_{den}$\\
    Calculate $\mathcal{L}_{C\!R\!T}$ by Eq.~(\ref{CRT part loss function})\\
    Generate segmentation-guided pseudo labels in $\mathcal{D_{T}}$\\
    Calculate $\mathcal{L}_{C\!D\!A}$ according to Eq.~(\ref{CDA loss function})\\
    Optimize $C(\cdot,\theta)$ by Eq.~(\ref{total loss function})
}
\end{algorithm}

\subsection{Crowd Density Alignment}


Crowd Density Alignment (CDA) is designed to regularize target-domain crowd density generation by its own crowd density distribution, instead of source-domain density distribution utilized in all the previous methods.


Specifically, we use soft crowd segmentation $\mathbf{C}_{seg}^S$ to generate probabilistic crowd distribution $P$ by normalization:
\begin{equation}
\label{pseudo label generation nomalize}
P = \mathbf{C}_{seg}^S\, / \sum_{h,w}\mathbf{C}_{seg}^S(h,w).
\end{equation}
$P$ follows a discrete bivariate distribution where we iteratively sample pseudo head positions $\mathcal{P} = \{(w_i, h_i)\,|\,i\!\in\![1,n]\}$. After sampling, we generate pseudo density labels as in the source domain by convolving each pseudo head point with a Gaussian kernel.

Following previous methods, we utilize adversarial training to regularize target-domain crowd density generation. Differently, we exploit segmentation-guided pseudo density labels as guidance, instead of source-domain density labels. The optimization objective is:
\begin{equation}
\label{CDA loss function}
\begin{split}
\mathcal{L}_{C\!D\!A} =\,\, & \underset{\theta_{G}}{{\rm min}} \, \underset{\theta_{D_m}}{{\rm max}} \, \mathbb{E}_{\mathbf{M}\!_{S\!P\!L}\!\sim\mathcal{D}\!_{S\!P\!L}} \, {\rm log} \, D_m(\mathbf{M}\!_{S\!P\!L})) \\
&+ \mathbb{E}_{\mathbf{x}_t\sim\mathcal{D_T}} \, {\rm log}(1 - D_m(G(\mathbf{x}_t))),
\end{split}
\end{equation}
where $D_m$ denotes crowd density discriminator. $\mathbf{M}\!_{S\!P\!L}$ and $\mathcal{D}\!_{S\!P\!L}$ represent segmentation-guided pseudo density maps and the corresponding domain respectively.
With the segmentation-guided pseudo labels, our method can directly constrain the target-domain crowd density generation by its own crowd density distribution. 

Note that the head-highlighted nature of soft crowd segmentation $\mathbf{C}_{seg}^S$ also benefits the generation of segmentation-guided pseudo labels considering the head-centric labeling mechanism of crowd counting.

\subsection{Network Optimization}
The training procedure of the proposed framework contains three major components: Supervised Learning (SL) $\mathcal{L}_{den}$, Crowd Region Transfer (CRT) $\mathcal{L}_{C\!RT}$, and Crowd Density Alignment (CDA) $\mathcal{L}_{C\!D\!A}$. With the above terms, the overall optimization objective writes as:
\begin{equation}\label{total loss function}
  \mathcal{L}_{total} = \mathcal{L}_{den} + \lambda_1\mathcal{L}_{C\!RT} + \lambda_2\mathcal{L}_{C\!D\!A},
\end{equation}
where $\lambda_1$ and $\lambda_2$ are factors to balance the three items.
Detailed optimization procedure is shown in Algorithm \ref{optimization procedure}.  

\section{Experiments}
\vspace{-4pt}
\subsection{Datasets and Adaptation Scenarios}
\vspace{-2pt}
\noindent \textbf{Datasets.} Six datasets are used in our experiments, i.e., GCC \cite{wang2019learning}, ShanghaiTech PartA (SHPartA) \cite{zhang2016single}, ShanghaiTech PartB (SHPartB) \cite{zhang2016single}, JHU-CROWD (JHUC)~\cite{sindagi2020jhu}, MALL~\cite{chen2012feature}, and UCSD~\cite{chan2008privacy}. Details are in the Appendix.

\noindent \textbf{Adaptation Scenarios.} (i) \textbf{Synthetic-to-Real} (GCC$\rightarrow$SHPartB, GCC$\rightarrow$SHPartA).
We employ the synthetic GCC as source domain and the training set of SHPartB or SHPartA as target domain.
(ii) \textbf{Fixed-to-Fickle} (SHPartB $\rightarrow$ SHPartA).
We utilize the training set of SHPartB (a fixed crowd scene) as source domain and the training set of SHPartA (various crowd scenes) as target domain.
(iii) \textbf{Normal-to-BadWeather} (SHPartA$\rightarrow$JHUC).
To simulate weather condition changes, we utilize the training set of SHPartA as source domain and the images with bad weather conditions in the training set of JHUC as target domain.

\vspace{-6pt}
\subsection{Ablation Studies}
\vspace{-2pt}
We conduct ablation studies in Synthetic-to-Real adaptation scenario to validate the effectiveness of the proposed modules, i.e., PCS, CRT, and CDA.


 \begin{table}[t]
	\centering
	\small
	\tabcolsep=0.10cm
	\caption{Ablation studies on Crowd Region Transfer (CRT) in the Synthetic-to-Real adaptation scenario.}
	\vspace{-4pt}
	\begin{tabular}{c||cc|cc}
	    \toprule[1pt]
       \multirow{2}{*}{Method}& \multicolumn{2}{c|}{GCC $\rightarrow$ SHPartB} & \multicolumn{2}{c}{GCC $\rightarrow$ SHPartA}\\
       &MAE&RMSE &MAE&RMSE\\
        \midrule
        \midrule
        Source only & 19.5 & 28.9 & 169.2 & 255.9 \cr
        CRT w/o PCS & 16.4 & 26.8 & 134.8 & 213.6 \cr
        CRT w/ BinarySeg. & 15.6 & 24.1 & 125.3 & 204.9\cr
        CRT w/ Hard Seg. & 15.0 & 23.8 & 122.5 & 203.2 \cr
        CRT w/ Soft Seg. & \textbf{14.7} & \textbf{23.5} & \textbf{117.4} & \textbf{201.6} \cr
        \bottomrule[1pt]
        \end{tabular}
	\label{CRT Table}
\end{table}

\begin{figure}[t]
 \centering
 \setlength{\tabcolsep}{0.7pt}
 \renewcommand{\arraystretch}{0.5}
  \begin{tabular}{cccc}
     \includegraphics[width=.24\linewidth, height=1.2cm]{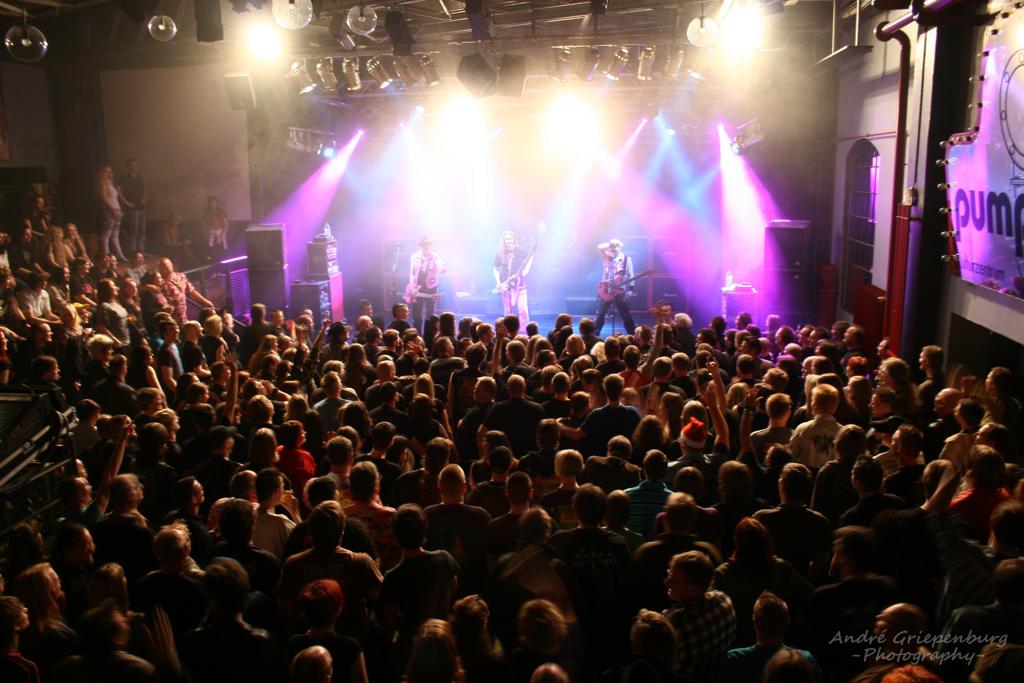}&
    \includegraphics[width=.24\linewidth, height=1.2cm]{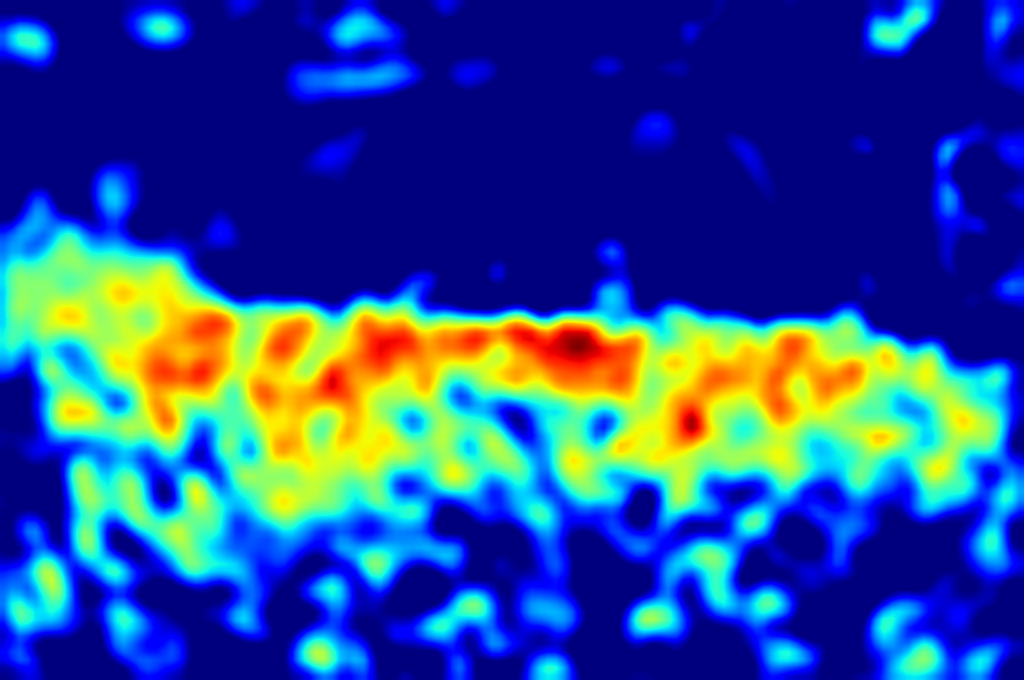}&
     \includegraphics[width=.24\linewidth, height=1.2cm]{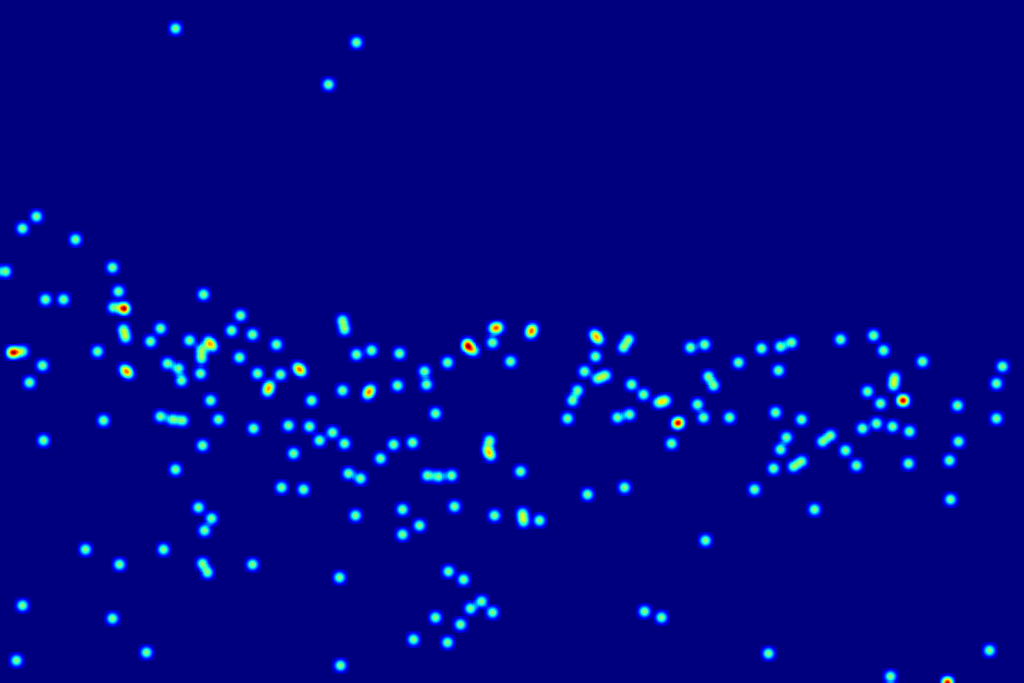}&
     \includegraphics[width=.24\linewidth, height=1.2cm]{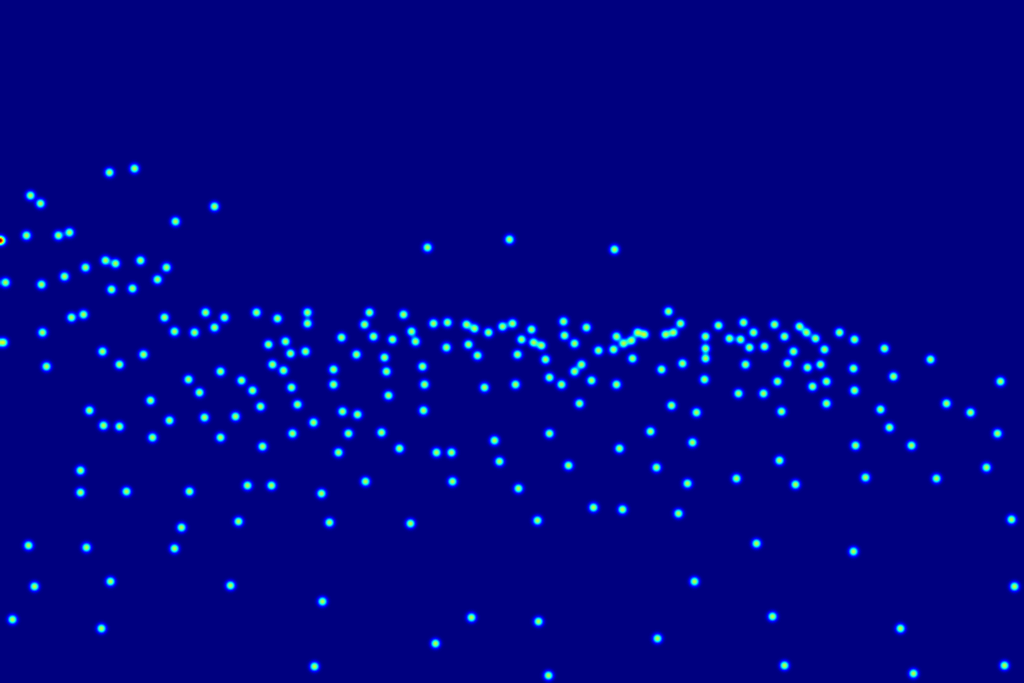}\\
     \includegraphics[width=.24\linewidth, height=1.2cm]{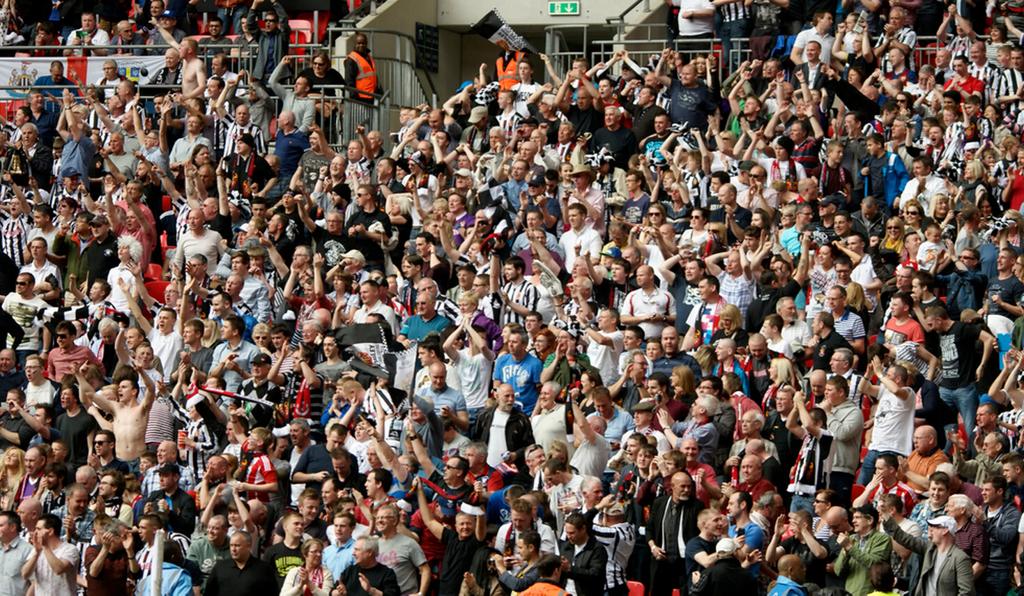}&
    \includegraphics[width=.24\linewidth, height=1.2cm]{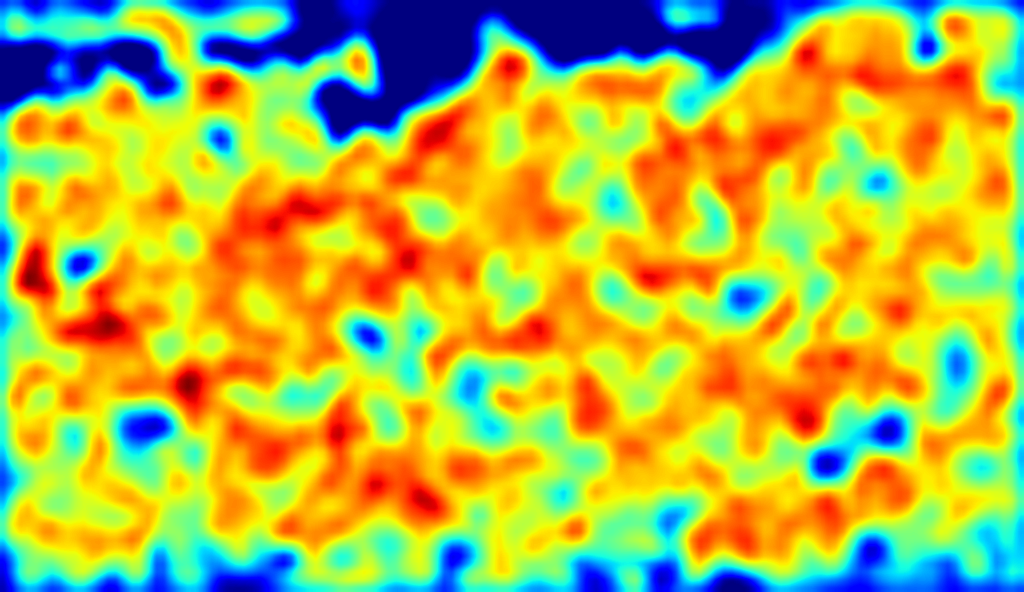}&
     \includegraphics[width=.24\linewidth, height=1.2cm]{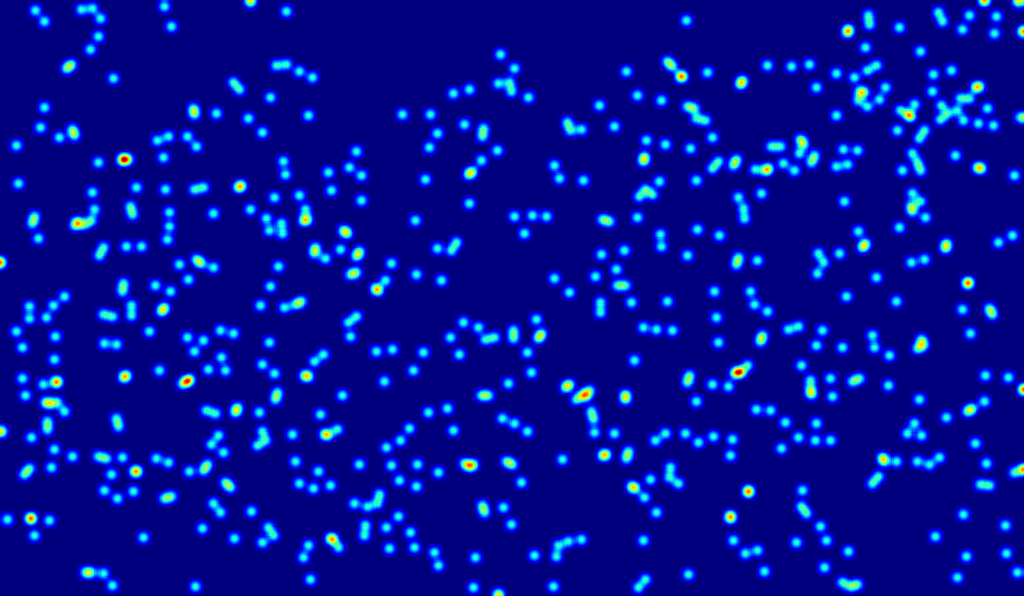}&
     \includegraphics[width=.24\linewidth, height=1.2cm]{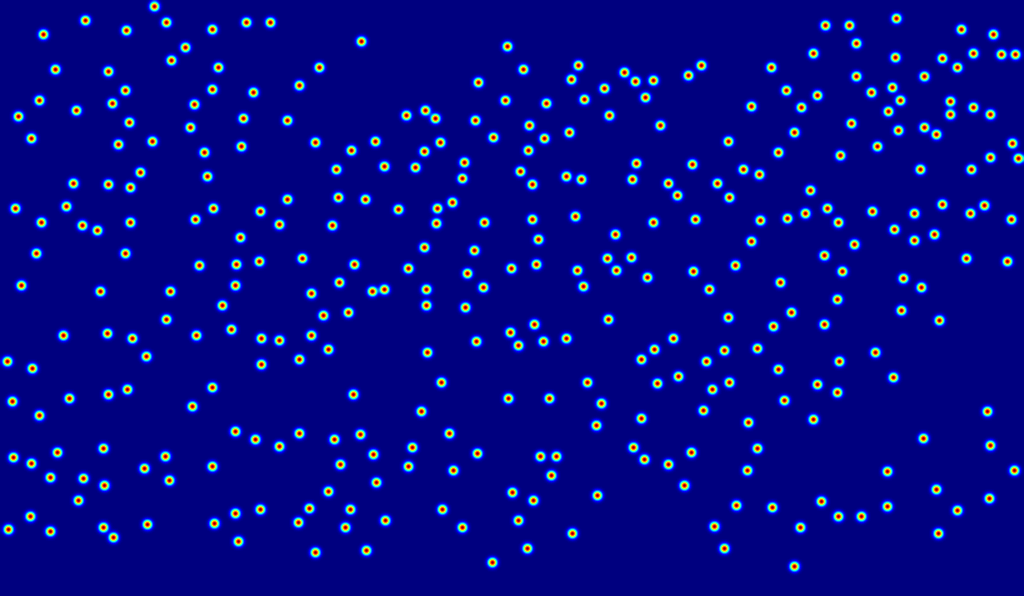}\\
      Crowd Image & PCS Seg. & CDA Label  & Ground Truth\\
  \end{tabular}
  \vspace{-2pt}
 \caption{Qualitative results of Point-derived Crowd Segmentation (PCS Seg.) and segmentation-guided pseudo label for Crowd Density Alignment (CDA label) in the target domains of Synthetic-to-Real adaptation scenario.}
 \vspace{-10pt}
 \label{PCS samples}
 \end{figure}

\begin{table*}[t]
	\centering
	\small
	\tabcolsep=0.100cm
	\setlength\arrayrulewidth{1.0pt}
	\caption{Comparison with state-of-the-art methods in the Synthetic-to-Real adaptation scenario. ``U'' and ``S'' denote unsupervised and semi-supervised domain adaptation methods, respectively. ``Gain'' denotes the relative gains of MSE/RMSE in comparison to the performance before adaptation.}
	\vspace{-6pt}
	\begin{tabular}{cc||ccc|ccc}
		\toprule
        \multicolumn{8}{c}{Synthetic $\rightarrow$ Real} \cr
        \midrule
		\multirow{2}{*}{Method}&\multirow{2}{*}{Type}&
		\multicolumn{3}{c|}{GCC $\rightarrow$ SHPartB}&\multicolumn{3}{c}{GCC $\rightarrow$ SHPartA}\\
		&&MAE $\downarrow$&RMSE $\downarrow$&Gain $\uparrow$&MAE $\downarrow$&RMSE $\downarrow$&Gains $\uparrow$\\
		\midrule
        \midrule
        NLT~\cite{wang2021neuron} & S & 10.8 & 18.3 & 46.2\%/37.3\% & 90.1 & 151.6 & 52.0\%/45.7\% \cr
        FSC~\cite{han2020focus} & S & 16.9 & 24.7 & 31.1\%/26.7\% & 129.3 & 187.6 & 32.2\%/37.0\% \cr
        \midrule
        CycleGAN~\cite{zhu2017unpaired} & U & 25.4 & 39.7 & -10.2\%/-23.6\% & 143.3 & 204.3 & 10.4\%/5.6\% \cr
        SE CycleGAN~\cite{wang2019learning} & U & 19.9 & 28.3 & 12.7\%/7.5\% & 123.4 & 193.4 & 22.8\%/10.6\% \cr
        FADA~\cite{gao2020feature} & U & 16.0 & 24.7 & 28.2\%/17.3\% & -- & -- & -- \cr
        ASNet~\cite{zou2021coarse} & U & 14.6 & 22.6 & -- & -- & -- & -- \cr
        Ours & U & \textbf{13.5} & \textbf{21.8} & \textbf{30.7\%}/\textbf{24.5\%} & \textbf{109.3} & \textbf{187.1} & \textbf{35.4\%}/\textbf{26.8\%} \cr
        \midrule
        Oracle & -- & 8.9 & 15.3 & -- & 67.5 & 112.1 & -- \cr
		\bottomrule
	\end{tabular}
	\label{SyntheticToReal Table}
\end{table*}

\begin{table}[t]
	\centering
	\small
	\tabcolsep=0.10cm
    \vspace{-8pt}
	\caption{Ablation studies on Crowd Density Alignment (CDA) in the Synthetic-to-Real adaptation scenario. CRT here utilizes soft crowd segmentation (``CRT w/ Soft Seg.'').}
	\vspace{-7pt}
	\begin{tabular}{c||cc|cc}
	    \toprule[1pt]
       \multirow{2}{*}{Method}& \multicolumn{2}{c|}{GCC $\rightarrow$ SHPartB} & \multicolumn{2}{c}{GCC $\rightarrow$ SHPartA}\\
       &MAE&RMSE &MAE&RMSE\\
        \midrule
        \midrule
        CRT & 14.7 & 23.5 & 117.4 & 201.6 \cr
        CRT + SL~\cite{gao2020feature} & 14.3 & 22.4 & 114.7 & 193.7 \cr
        CRT + CDA & \textbf{13.5} & \textbf{21.8} & \textbf{109.3} & \textbf{187.1} \cr
		\bottomrule[1pt]
	\end{tabular}
  \vspace{-8.67pt}
	\label{CDA Table}
\end{table}

\noindent \textbf{Effectiveness of PCS.}
We evaluate PCS by testing how much the derived crowd segmentation covers annotated human heads.
The percentages of coverage in Synthetic-to-Real adaptation scenario are 98.5, 93.6, and 95.2 for GCC (source domain), SHPartA (target domain), and SHPartB (target domain) datasets, respectively. This indicates that point-derived crowd segmentation can cover almost all human heads in both domains.
Qualitative results of PCS are shown in Fig.~\ref{PCS samples}.

 
 \noindent \textbf{Effectiveness of CRT.}
 To evaluate the effective of CRT, we introduce several comparison variants as follows. 
 ``Source only'' denotes crowd counter trained on source domain only.
``CRT w/o PCS'' transfers features across domains in a holistic manner.
``CRT w/ Hard Seg.'', ``CRT w/ Soft Seg.'', and ``CRT w/ BinarySeg.'' denote CRT with hard crowd segmentation, soft crowd segmentation, and binarizing Gaussian-blurred density maps~\cite{liu2020semi}.
  
As shown in Table~\ref{CRT Table}, compared to ``Source Only'', ``CRT w/o PCS'' can improve the adaptation performance to some extent. 
 ``CRT w/ BinarySeg.'',  ``CRT w/ Hard Seg.'', and ``CRT w/ Soft Seg.'' achieve lower counting errors compared to ``CRT w/o PCS'' no matter what kind of crowd segmentation is leveraged. This indicates background features alignment across domains incurs an adverse effect during domain adaptation.
  ``CRT w/ Soft Seg.'' is better than ``CRT w/ Hard Seg.'', which demonstrates the effectiveness of enhanced head features brought by head-highlighted soft crowd segmentation. 

\noindent \textbf{Effectiveness of CDA.}
As shown in Table~\ref{CDA Table}, ``CDA'' outperforms ``SL'' (Source-domain density Labels)~\cite{gao2020feature} consistently, which demonstrates the superiority of the segmentation-guided density alignment mechanism.
Qualitative results of segmentation-guided pseudo labels are in Fig.~\ref{PCS samples}.

 \vspace{-8pt}
\subsection{Comparison to state-of-the-art methods}
 \vspace{-4pt}

\noindent \textbf{Synthetic-to-Real.} As shown in Table~\ref{SyntheticToReal Table}, our method can achieve the lowest counting errors and the highest relative gains compared to all the unsupervised counterparts.
Although we do not leverage extra annotations, our method can still outperform FSC~\cite{han2020focus}.
To be comparable with NLT~\cite{wang2021neuron}, we also introduce 10\% labeled target data.
The performance of our method in terms of MAE/RMSE is enhanced to 10.2/17.5 (SHPartB) and 82.4/136.6 (SHPartA), respectively, which are better than NLT~\cite{wang2021neuron}.

\noindent \textbf{Fixed-to-Fickle \& Normal-to-BadWeather.}
The two adaptation scenarios are discussed for the first time in the literature. Due to limited space, we show the results in the Appendix.

\noindent \textbf{Others.}
To conduct more comparisons with state-of-the-art methods, we follow some other shared settings, i.e., SHPartA$\rightarrow$SHPartB, MALL$\rightarrow$UCSD, and UCSD$\rightarrow$MALL. As can be seen in Table~\ref{New Table}, our method can outperform state-of-the-art methods in different adaptation scenarios.

\begin{table}[t]
	\centering
	\scriptsize
	\tabcolsep=0.100cm
	\setlength\arrayrulewidth{1.0pt}
    \vspace{-6pt}
	\caption{Comparisons with state-of-the-art methods on some shared settings, i.e., SHPartA$\rightarrow$SHPartB, MALL$\rightarrow$UCSD, and UCSD$\rightarrow$MALL.}
	\vspace{-2pt}
	\begin{tabular}{cc||cc|cc|cc}
		\toprule
		\multirow{2}{*}{Method}&\multirow{2}{*}{Type}&
		\multicolumn{2}{c|}{SHPartA $\rightarrow$ SHPartB }&\multicolumn{2}{c|}{MALL $\rightarrow$ UCSD}&\multicolumn{2}{c}{UCSD $\rightarrow$ MALL}\\
		&&MAE $\downarrow$&RMSE $\downarrow$&MAE $\downarrow$&RMSE $\downarrow$&MAE $\downarrow$&RMSE $\downarrow$\\
		\midrule
        \midrule
        DACC~\cite{hossain2020domain} & U & -- & -- & 2.52 & 3.38 & 2.93 & 3.65 \cr
        ASNet~\cite{zou2021coarse} & U & 13.59 & 23.15 & -- & -- & 2.76 & 3.55 \cr
        Ours & U & \textbf{12.84} & \textbf{21.92} & \textbf{2.39} & \textbf{3.26} & \textbf{2.68} & \textbf{3.50} \cr
		\bottomrule
	\end{tabular}
	\label{New Table}
\end{table}

\begin{figure*}[t]
 \centering
 \setlength{\tabcolsep}{0.3pt}
 \renewcommand{\arraystretch}{0.3}
  \begin{tabular}{ccccc}
     \includegraphics[width=.18\linewidth, height=1.8cm]{./figures/27_img.png}&
     \includegraphics[width=.18\linewidth, height=1.8cm]{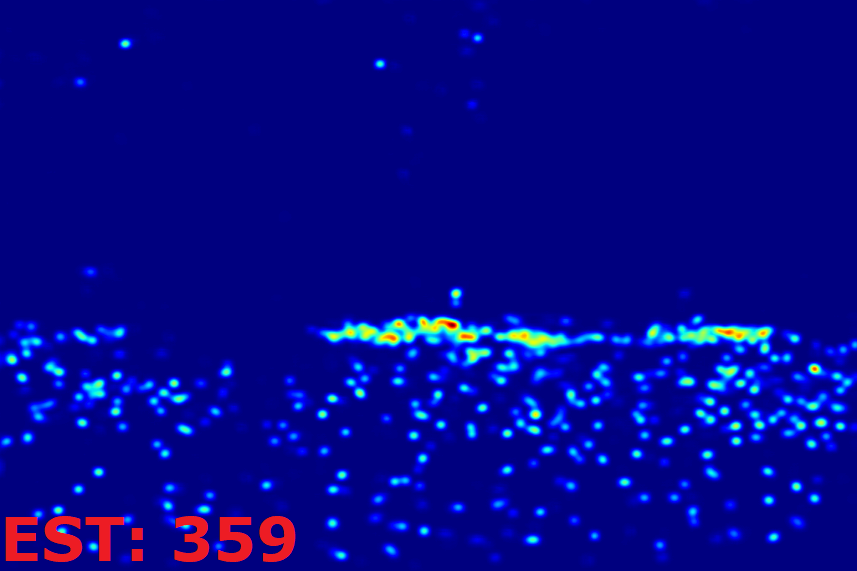}&
     \includegraphics[width=.18\linewidth, height=1.8cm]{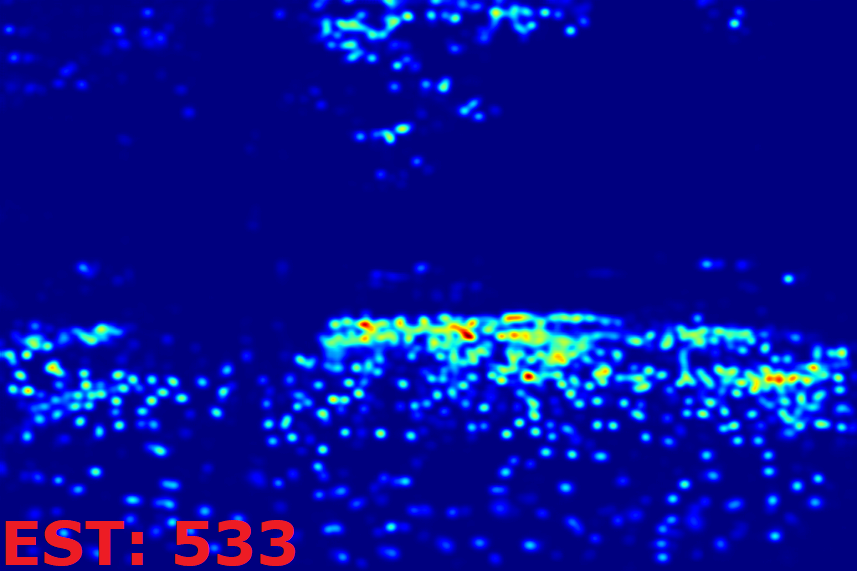}&
     \includegraphics[width=.18\linewidth, height=1.8cm]{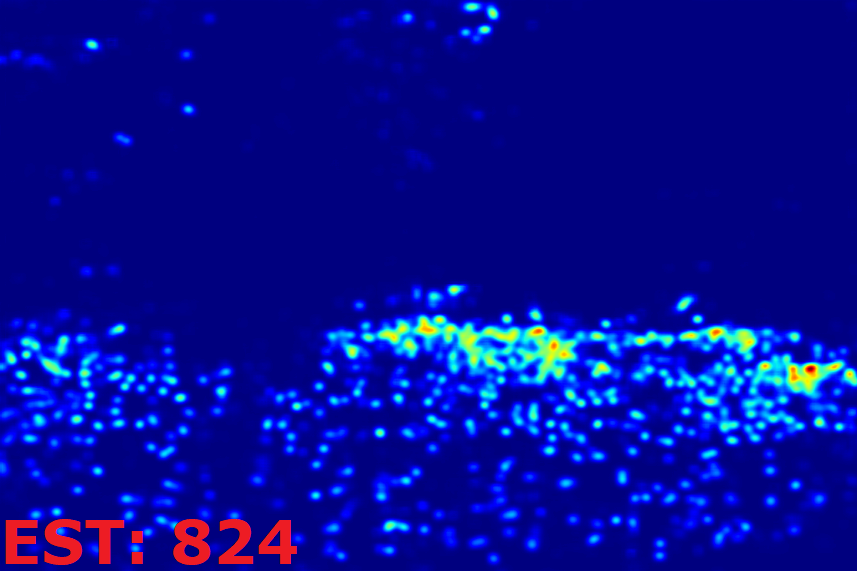}&
     \includegraphics[width=.18\linewidth, height=1.8cm]{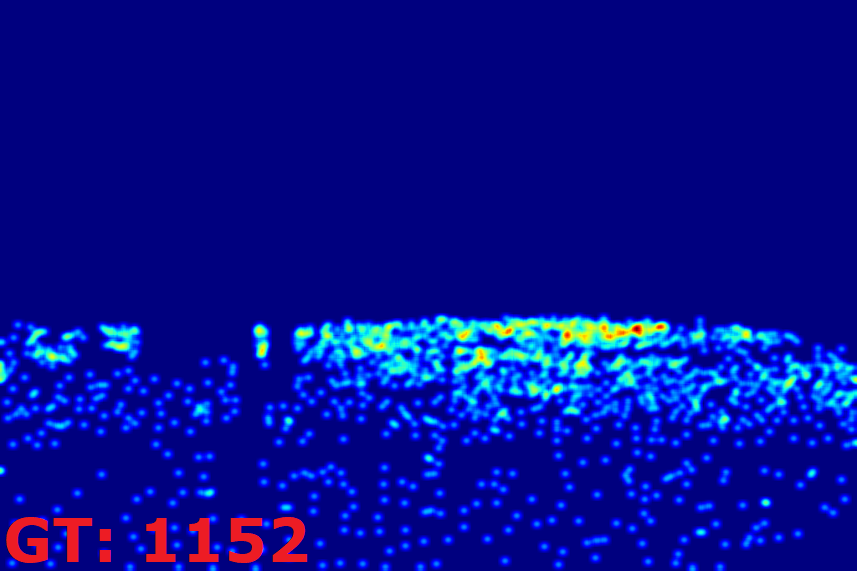}\\
     \includegraphics[width=.18\linewidth, height=1.8cm]{./figures/60_img.png}&
     \includegraphics[width=.18\linewidth, height=1.8cm]{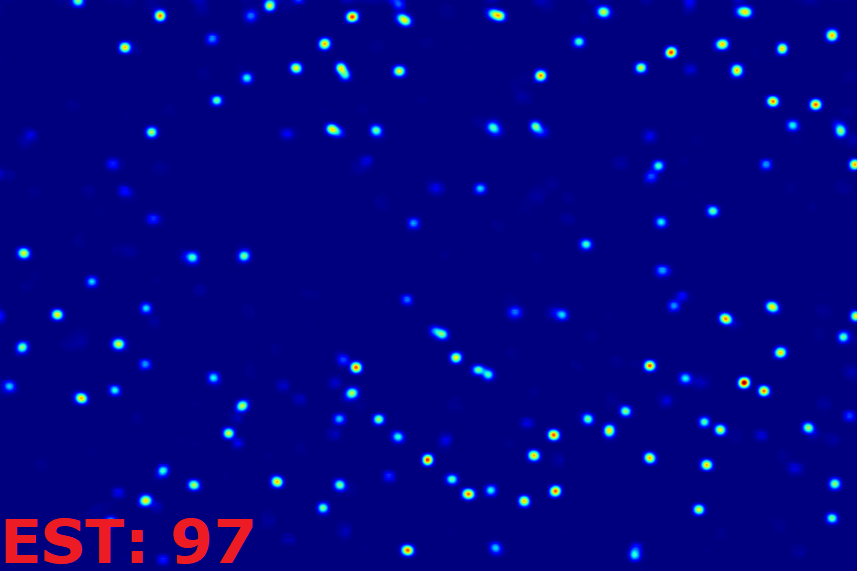}&
     \includegraphics[width=.18\linewidth, height=1.8cm]{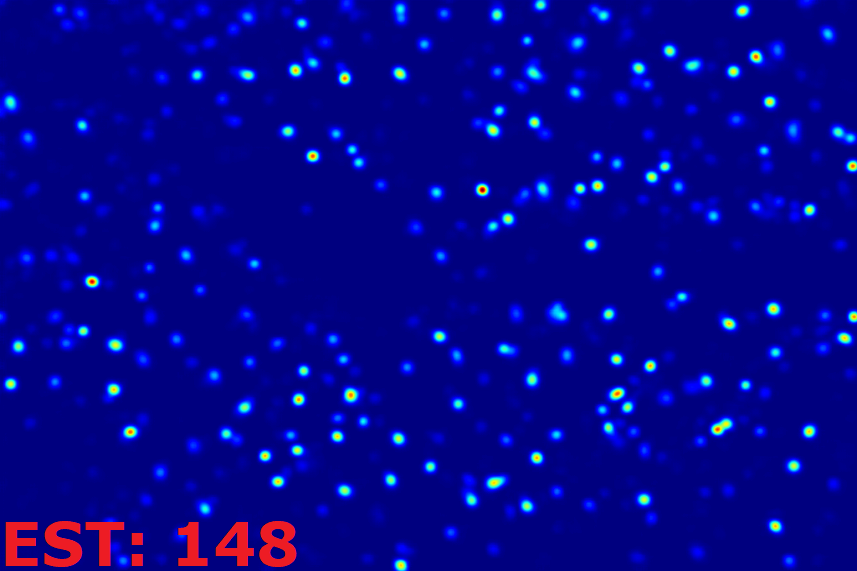}&
     \includegraphics[width=.18\linewidth, height=1.8cm]{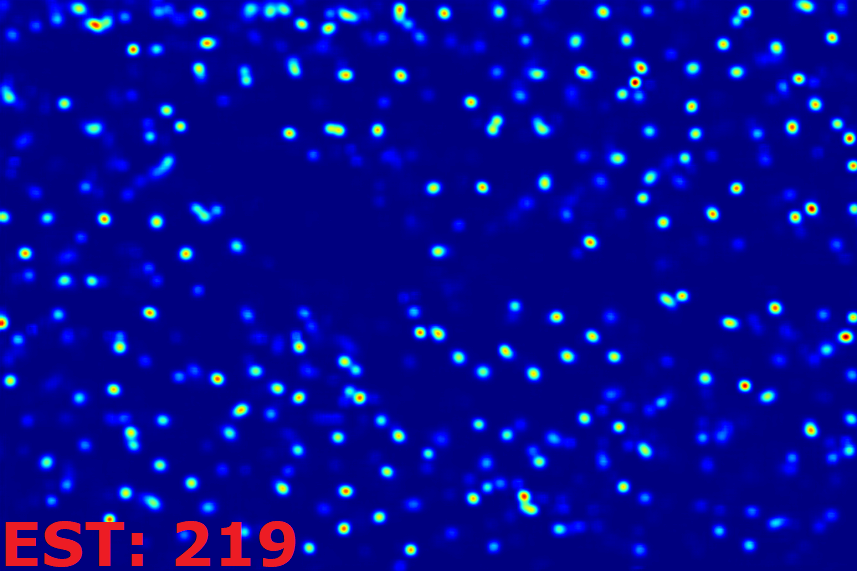}&
     \includegraphics[width=.18\linewidth, height=1.8cm]{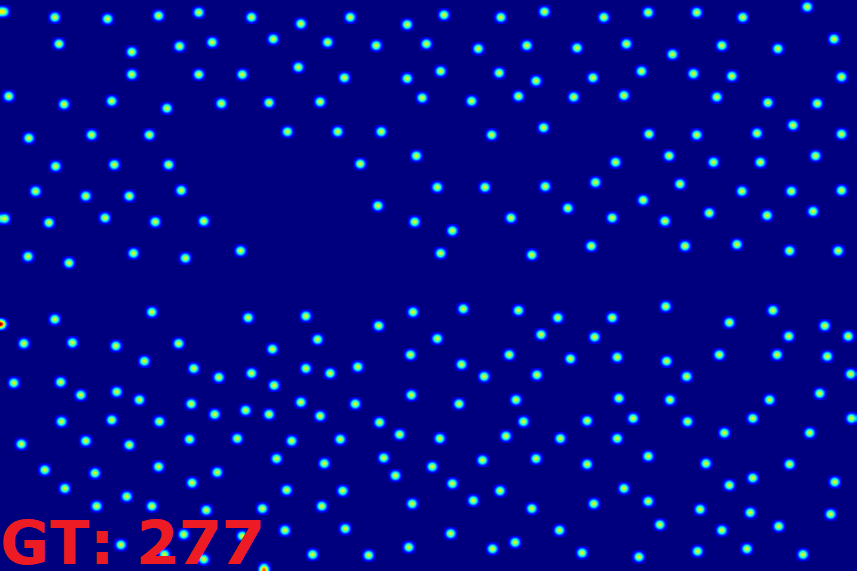}\\
     \includegraphics[width=.185\linewidth, height=2.1cm]{./figures/81_img.png}&
     \includegraphics[width=.185\linewidth, height=2.1cm]{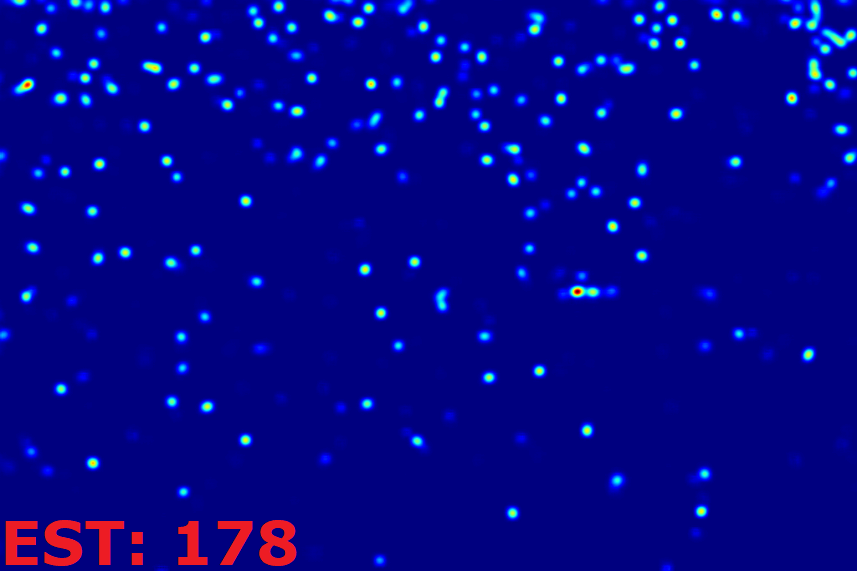}&
     \includegraphics[width=.185\linewidth, height=2.1cm]{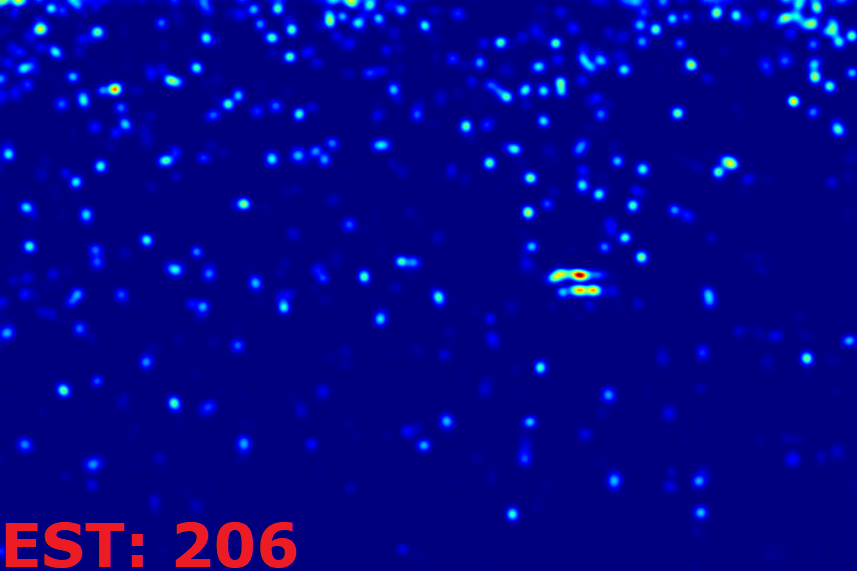}&
     \includegraphics[width=.185\linewidth, height=2.1cm]{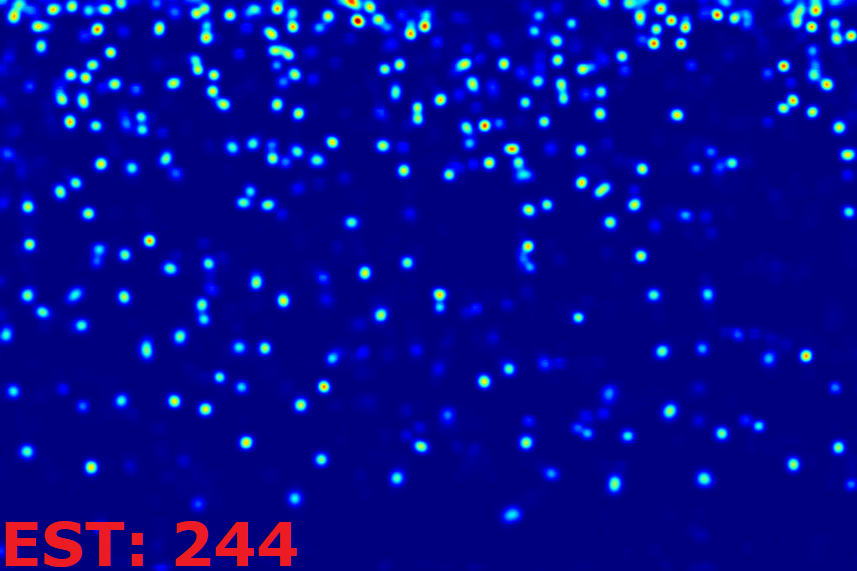}&
     \includegraphics[width=.185\linewidth, height=2.1cm]{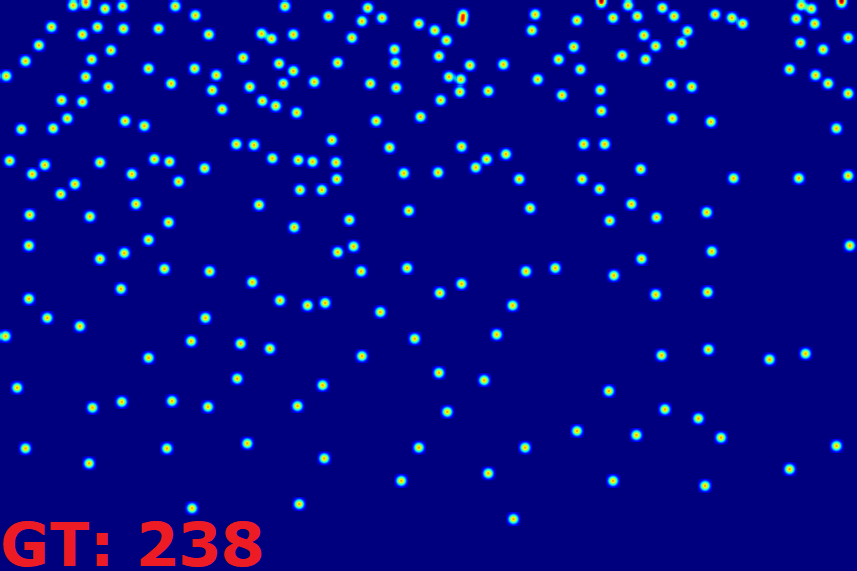}\\
     Input Image & Source Only & Ours w/o PCS & Ours & Ground Truth\\
  \end{tabular}
  \vspace{-4pt}
 \caption{Qualitative results of the estimated density maps in the Synthetic-to-Real adaptation scenario. Note that ``Ours w/o PCS'' also means our method without the CRT and CDA modules as PCS is the base of them.}
 \label{QualitativeCP}
 \end{figure*}

\subsection{Qualitative Results}
Qualitative results of the estimated density maps can be seen in Fig.~\ref{QualitativeCP}.
Due to the domain shift problem, the ``Source Only'' model simply detects some salient individuals in the crowd. 
From ``Source Only'' to ``Ours w/o PCS'', we can observe that the ``Ours w/o PCS'' model can increase true positives to some extent, but also incurs some false positives in the background areas due to the misalignment between crowd and background. Differently, our method can consistently estimates more accurate crowd densities and suppresses the occurrence of false positives thanks to the proposed crowd-aware domain adaptation method.

\vspace{-3pt}
\section{Conclusion}
\vspace{-3pt}
In this paper, we propose to treat crowd and background differently and design a Crowd-aware domain Adaptation framework for Crowd Counting (CACC).
Specifically, we learn crowd segmentation from pixel-level crowd counting annotations.
Based on the derived segments, we design two crowd-aware adaptation modules, i.e., Crowd Region Transfer (CRT) and Crowd Density Alignment (CDA).
Extensive experiments in multiple cross-domain scenarios demonstrate the superiority of the proposed method.

\bibliographystyle{IEEEbib}
\bibliography{2023template}

\section{Appendix}
\subsection{More Related Work}
\subsubsection{Crowd Counting}
Early works for crowd counting are mainly based on hand-crafted features (e.g., SIFT, Fourier Analysis, HOG) to estimate crowd counts by either regression~\cite{chen2012feature, idrees2013multi, lempitsky2010learning} or object detection~\cite{lin2010shape, wang2011automatic, wu2007detection}.
Various CNN-based methods have advanced the performance of crowd counting.
Most of them are dedicated to handle various challenges of crowd counting in an supervised manner, e.g., large scale variations~\cite{zhang2016single, cao2018scale, li2018csrnet, liu2019adcrowdnet, sam2017switching, liu2019context, jiang2019crowd, luo2020hybrid, jiang2020attention}, hand-crafted gaussian kernels~\cite{wan2019adaptive, ma2019bayesian, liu2020weighing}, uncertainty~\cite{ranjan2020uncertainty, oh2020crowd}, enhancing crowd features~\cite{sindagi2019ha, miao2020shallow, zhang2019relational, zhang2019attentional}, extra constraints~\cite{lian2019density, shi2019revisiting, yan2019perspective}, etc.
Besides the supervised methods, several approaches focus on relieving the labeling burdensome. They can be broadly categorized into semi-supervised methods~\cite{zhou2018crowd, reddy2020few, sindagi2020learning, zhao2020active, liu2020semi, liu2022reducing}, weakly-supervised methods~\cite{yang2020weakly, xiong2022glance}, self-supervised methods~\cite{liu2018leveraging, liu2019exploiting} and unsupervised methods~\cite{elassal2016unsupervised, sam2019almost}.

These generic crowd counters can achieve promising performance in public datasets, whereas they do not focus on solving the domain shift problem, which hurts their generalization performance in real-world application scenarios.

\subsubsection{Domain Adaptation}
Lots of domain adaptation methods dedicate to reducing domain discrepancies by learning domain-invariant feature representations. Methods along this line can be generally categorized into two types: criterion-based methods~\cite{long2017deep, long2018transferable, sun2016deep, zellinger2017central, rozantsev2018residual, lee2019sliced, chen2020homm} and adversarial learning-based methods~\cite{ganin2015unsupervised, ganin2016domain, tzeng2017adversarial, chen2018re, zhang2018collaborative, volpi2018adversarial, chen2019progressive, zhang2019domain, lee2019drop, chen2020adversarial, xu2020adversarial}.
The former aligns feature distributions between different domains by minimizing some statistics, such as Maximum Mean Discrepancy~\cite{long2017deep}, Correlation Alignment~\cite{sun2016deep}, Wasserstein distance~\cite{lee2019sliced}, and HoMM~\cite{chen2020homm}.
The latter introduces a domain discriminator to classify feature representations, while adversarially confuses the discriminator by constructing a minimax game with the feature extractor.
These methods have been widely studied and achieved superior performance in image classification~\cite{ganin2015unsupervised, saito2017asymmetric, ding2018graph, kang2019contrastive}, semantic segmentation~\cite{sankaranarayanan2018learning, tsai2018learning, luo2019taking, du2019ssf, yang2020adversarial, wang2020differential, pan2020unsupervised}, object detection~\cite{chen2018domain, kim2019diversify, saito2019strong, zhu2019adapting, cai2019exploring, zheng2020cross, sindagi2020prior}.

However, domain adaptation for crowd counting is less studied, and existing generic methods cannot easily adapt to crowd counting due to its special labeling mechanism and diverse backgrounds in crowd scenes.

\subsection{More Network Details}

\begin{table}[t]
	\centering
	\small
	\tabcolsep=0.100cm
	\setlength\arrayrulewidth{1.0pt}
	\caption{Architecture of crowd counter.}
	\begin{tabular}{c}
		\toprule
		\textbf{VGG16 Backbone} \cr
		Conv1: [K(3,3)-C64-S1-R] \cr
		... \cr
		Conv10: [K(3,3)-C512-S1-R] \cr
        \midrule
        \textbf{Deconvolution Block} \cr
        Conv11: [K(3,3)-C64-S1-R]; Deconv1: [K(2,2)-C64-S2-R]\cr
        Conv12: [K(3,3)-C32-S1-R]; Deconv2: [K(2,2)-C32-S2-R] \cr
        Conv13: [K(3,3)-C16-S1-R]; Deconv3: [K(2,2)-C16-S2-R] \cr
        \midrule
        \textbf{Density Regression Layer} \cr
        Conv14: [K(3,3)-C16-S1-R] \cr
        Conv15: [K(3,3)-C1-S1-R] \cr
		\bottomrule
	\end{tabular}
	\label{Crowd Counting Network}
\end{table}

\begin{table}[t]
	\centering
	\small
	\tabcolsep=0.100cm
	\setlength\arrayrulewidth{1.0pt}
	\caption{Architecture of Point-derived Crowd Segmentation (PCS) network.}
	\begin{tabular}{c}
		\toprule
		\textbf{Feature Extractor} \cr
		Conv1: [K(3,3)-C16-S1-R]; Conv2: [K(3,3)-C16-S1-R] \cr
		MaxPool1: [K(2,2)-C16-S2] \cr
		Conv3: [K(3,3)-C32-S1-R]; Conv4: [K(3,3)-C32-S1-R] \cr
		MaxPool2: [K(2,2)-C32-S2] \cr
		Conv5: [K(3,3)-C32-S1-R]; Conv6: [K(3,3)-C32-S1-R] \cr
		MaxPool3: [K(2,2)-C32-S2] \cr
        Conv7: [K(3,3)-C2-S1-R] \cr
        \midrule
        \textbf{2DAvgPool \& Softmax} \cr
		\bottomrule
	\end{tabular}
	\label{point-derived segmentation network}
\end{table}

\subsubsection{Architecture of Crowd Counter}
Most crowd counting networks employ density maps as the intermediate output for better supervision. They are typically generated by convolving each annotated head point with a Gaussian kernel~\cite{zhang2016single}:
\begin{equation}
\label{density map generation}
    \mathcal{D}(\mathbf{z})=\sum_{k=1}^{N}\delta(\mathbf{z}-\mathbf{z}_{k}) * G_{\sigma_{k}}(\mathbf{z}),
\end{equation}
where $\mathbf{z}$ and $\mathbf{z}_{k}$ denote each pixel and the $k$-th annotated point (total $N$ points) in a crowd image $\mathbf{x}$. $G_{\sigma_{k}}$ is a 2D Gaussian kernel with a bandwidth $\sigma_{k}$.
Following previous works~\cite{gao2020feature, wang2019learning, han2020focus}, we employ a simple and universal crowd counter without specialized techniques to verify the general effectiveness of the proposed domain adaptation method.
Specifically, we extract the first ten convolutional layers of VGG16~\cite{simonyan2014very} with three maxpooling layers as the backbone network.
After the backbone network, we introduce several deconvolutional layers to generate high-resolution density maps. Detailed network architecture of crowd counter is in Table~\ref{Crowd Counting Network}. For example, ``K(3,3)-C64-S1-R'' represents the Convolution or Deconvolution layer with kernel size of $3 \times 3$, 64 output channels, stride size of 1, and ReLU activation function.

To measure distance between ground truth and estimated density map, we adopt the widely-used pixel-wise Euclidean loss which can be formulated as:
\begin{equation}
\label{counting loss function}
    \mathcal{L}_{den}(\mathbf{x}) = \frac{1}{2M} \left\| \mathbf{D}^{est}(\mathbf{x}) - \mathbf{D}^{GT}(\mathbf{x}) \right\|_{2}^{2}
\end{equation}
where $M$ is the number of pixels in the input image $\mathbf{x}$. $\mathbf{D}^{est}(\mathbf{x})$ and $\mathbf{D}^{GT}(\mathbf{x})$ represent the estimated and ground truth density maps, respectively.

\subsubsection{Architecture of Point-derived Crowd Segmentation}
Detailed network architecture of the Point-derived Crowd Segmentation (PCS) network is in Table~\ref{point-derived segmentation network}.

\begin{table}[t]
	\centering
	\small
	\tabcolsep=0.10cm
    \vspace{-2pt}
	\caption{More ablation studies on Crowd Density Alignment (CDA) in the Synthetic-to-Real adaptation scenario.}
	\vspace{-2pt}
	\begin{tabular}{c||cc|cc}
	    \toprule[1pt]
       \multirow{2}{*}{Method}& \multicolumn{2}{c|}{GCC $\rightarrow$ SHPartB} & \multicolumn{2}{c}{GCC $\rightarrow$ SHPartA}\\
       &MAE&RMSE &MAE&RMSE\\
        \midrule
        \midrule
        Source only  & 19.5 & 28.9 & 169.2 & 255.9 \cr
        SL~\cite{gao2020feature} & 18.6 & 28.2 & 165.4 & 248.5 \cr
        CDA & \textbf{16.9} & \textbf{26.7} & \textbf{160.5} & \textbf{239.6} \cr
		\bottomrule[1pt]
	\end{tabular}
	\label{CDA Table}
\end{table}

\subsection{More Experiment Details}
\subsubsection{Implementation Details}
In all experiments, we set the batch size as 2, i.e., one image per domain.
We adopt random cropping and horizontal flipping for data augmentation.
Adam optimizer~\cite{kingma2014adam} is utilized to optimize the networks with the learning rate of the crowd counter and all classifiers initialized as $\mathrm{10}^{-\mathrm{4}}$ and $\mathrm{10}^{-\mathrm{5}}$, respectively.
$\lambda_1$ and $\lambda_2$ in Eq. (6) are set to 1 and 0.3, respectively via cross validation.
Following \cite{wang2019learning}, scene regularization is utilized to select synthetic images from GCC to facilitate adaptation.
We adopt three domain classifiers for multi-scale features extracted after each pooling layer in $G(\cdot)$ of Eq.~(4).
The training and evaluation are achieved on 2 NVIDIA GTX 2080Ti GPU.
Evaluation metrics are MAE and RMSE~\cite{zhang2016single}.

\subsubsection{Datasets}
Six datasets are utilized in our experiments.
(i) \textbf{GCC} \cite{wang2019learning} is a synthetic dataset containing 15,212 images with resolution of 1080 $\times$ 1920, which are rendered by GTA5 and captured by 400 surveillance cameras in a fictional city. 
(ii) \textbf{SHPartA} \cite{zhang2016single} is randomly crawled from the Internet with various crowd scenes containing 482 images, in which 300 images for training and 182 images for testing.
(iii) \textbf{SHPartB} \cite{zhang2016single} is collected from the busy streets of metropolitan areas in Shanghai consisting of 716 images, in which 400 images for training and the remaining for testing. Compared to SHPartA, SHPartB has relatively fixed camera perspectives and crowd scenes.
(iv) \textbf{JHU-CROWD (JHUC)} \cite{sindagi2020jhu} is a large-scale dataset proposed recently, which contains 4,372 images. Images are collected under a variety of scenes and environmental conditions, and annotations include head positions, approximate sizes, blur-level, occlusion-level, weather-labels, \emph{etc}. 
(v) \textbf{MALL} \cite{chen2012feature} is captured in a shopping mall by a fixed surveillance camera. The dataset consists of 2,000 frames in which the first 800 frames for training and the remaining for testing.
(vi) \textbf{UCSD} \cite{chan2008privacy} is collected by a fixed video camera besides a pedestrian walkway. The datasets contains 2,000 frames in which the training set captures 601 to 1,400 and the testing set owns the remaining. Region-of-interest (ROI) and perspective map are provided.

\begin{table}[t]
	\centering
	\small
	\tabcolsep=0.15cm
	\caption{Results of the Fixed-to-Fickle adaptation.}
	\vspace{-2pt}
	\begin{tabular}{c||ccc}
		\toprule[1pt]
        \multicolumn{4}{c}{Fixed $\rightarrow$ Fickle (SHPartB $\rightarrow$ SHPartA)} \cr
        \midrule
		Method&MAE $\downarrow$&RMSE $\downarrow$&Gain $\uparrow$\\
        \midrule
        \midrule
        Source only & 194.0 & 298.4 & -- \cr
        CRT w/o PCS & 153.2 & 247.5 & 21.0\%/17.1\% \cr
        CRT & 123.3 & 204.6 & 36.4\%/31.4\% \cr
        CRT + CDA (Ours) & \textbf{115.6} & \textbf{199.5} & \textbf{40.4\%}/\textbf{33.1\%} \cr
        \midrule
        Oracle & 67.5  & 112.1 & -- \cr
		\bottomrule[1pt]
	\end{tabular}
	\label{FixedToFickle Table}
\end{table}

\subsubsection{More Quantitative Results}

\noindent \textbf{Effectiveness of CDA.}
To further verify the effectiveness of Crowd Density Alignment (CDA), we conduct more ablation studies without based on the proposed Crowd Region Transfer (CFA). The comparison results are in Table~\ref{CDA Table}. We can see that “CDA” outper- forms “SL”, which demonstrates consistent superiority of the proposed segmentation- guided density alignment mechanism.

\noindent \textbf{Fixed-to-Fickle \& Normal-to-BadWeather.}
The two adaptation scenarios are discussed for the first time in the literature. However, they are also very important adaptation scenarios considering various crowd scenes and weather conditions in real-world applications. Results of different variants of our method in the two scenarios are summarized in Table~\ref{FixedToFickle Table}.
As can be seen, the proposed PCS, CRT, and CDA modules can progressively improve the counting accuracies in both adaptation scenarios, which confirms the effectiveness of the proposed crowd-aware domain adaptation mechanism in multiple real-world adaptation scenarios.

\begin{table}[t]
	\centering
	\small
	\tabcolsep=0.15cm
	\caption{Results of the Normal-to-BadWeather adaptation.}
	\vspace{-2pt}
	\begin{tabular}{c||ccc}
		\toprule[1pt]
        \multicolumn{4}{c}{Normal $\rightarrow$ BadWeather (SHPartA $\rightarrow$ JHUC)} \cr
        \midrule
		Method&MAE $\downarrow$&RMSE $\downarrow$&Gain $\uparrow$\\
        \midrule
        \midrule
        Source only & 208.5 & 535.6 & -- \cr
        CRT w/o PCS & 173.6 & 437.2 & 16.7\%/18.3\% \cr
        CRT w/ PCS & 159.5 & 394.7 & 23.5\%/26.3\%\cr
        CRT w/ PCS + CDA (full) & \textbf{153.2} & \textbf{384.0} & \textbf{26.5\%}/\textbf{28.3\%}\cr
        \midrule
        Oracle & 80.4 & 215.3 & -- \cr
		\bottomrule[1pt]
	\end{tabular}
	\label{NormalToBadWeather Table}
\end{table}


\end{document}